%% file: acl_latex.tex
\documentclass[11pt]{article}

\usepackage[final]{acl}

\usepackage{times}
\usepackage{latexsym}

\usepackage[T1]{fontenc}

\usepackage[utf8]{inputenc}

\usepackage{microtype}

\usepackage{inconsolata}

\usepackage{graphicx}

\usepackage{enumitem}
\usepackage{amsmath}
\usepackage{todonotes}
\usepackage{booktabs}
\usepackage{xcolor}
\usepackage{multirow} 
\usepackage{amssymb}
\usepackage{marvosym}
\usepackage[table]{xcolor}
\usepackage{tcolorbox}
\tcbuselibrary{breakable}
\usepackage{tabularx}
\usepackage{caption}
\usepackage{hyperref}

\title{Meta-Task: Turning Terminal Task Synthesis into a Terminal Task for Scalable Agent Training}

\author{
    \textbf{Zhihong Pan\textsuperscript{1}\dag},
    \textbf{Jiyuan He\textsuperscript{2}},
    \textbf{Kai Zhang\textsuperscript{1}\Letter},
    \textbf{Yupeng Han\textsuperscript{1}},
\\
    \textbf{Ze Liu\textsuperscript{1}},
    \textbf{Yuze Zhao\textsuperscript{1}},
    \textbf{Yongcong Ye\textsuperscript{1}},
    \textbf{Zhaohua Yang\textsuperscript{2}}
\\
\\
    \textsuperscript{1}State Key Laboratory of Cognitive Intelligence, \\
      University of Science and Technology of China,
    \textsuperscript{2}Meituan
\\
\small{
    \dag~Work done while interning at Meituan. \quad
    \Letter~Corresponding author.
}
\\
\small{
    \textbf{Correspondence:}
    \href{mailto:panzh@mail.ustc.edu.cn}{panzh@mail.ustc.edu.cn},
    \href{mailto:hejiyuan@meituan.com}{hejiyuan@meituan.com},
    \href{mailto:kkzhang08@ustc.edu.cn}{kkzhang08@ustc.edu.cn}
}
}

\begin{document}
\maketitle

\input{sections/abstract}

\input{sections/introduction}

\input{sections/method}

\input{sections/experiment}

\input{sections/related_work}
\input{sections/conclusion}
\input{sections/limitation}



\input{acl_latex.bbl}
\appendix
\input{sections/appendix}

\end{document}

%% file: sections/abstract.tex
\begin{abstract}
Training terminal agents at scale requires diverse, verifiable terminal tasks and high-quality interaction trajectories, yet acquiring such data remains a significant challenge. Existing synthesis methods face two key limitations: (1) weak reliability caused by the disconnect between task generation and real execution, and (2) limited diversity and scalability due to dependence on existing repositories. We propose \textbf{Meta-Task}, a framework that redefines terminal task synthesis as a Terminal-Bench-format task itself: an agent operates within a real container environment to iteratively generate, execute, and verify tasks, so that synthesized components are checked for internal consistency and executability within the generation loop itself. Building upon this, we decouple the target task requirements along multiple dimensions, introduce a multi-phase mechanism that dynamically designs novel task specifications before producing the actual tasks, and incorporate optional external material support to enhance diversity and realism. We additionally apply LLM-as-Judge filtering to ensure the quality of the final training data. Experiments on Terminal-Bench 2.0 show that fine-tuning on only 3,221 Meta-Task synthesized trajectories achieves \textbf{22.5\%} and \textbf{31.8\%} Avg Pass@1 for Qwen3-14B and Qwen3-32B respectively, outperforming concurrent approaches with significantly less training data.
\end{abstract}

%% file: sections/introduction.tex
\section{Introduction}

As the agentic capabilities of large language models (LLMs) continue to evolve, terminal interaction has emerged as one of the most critical competencies \cite{jimenez2024swe,merrill2026terminal}. The practical effectiveness of CLI-based agent tools such as Claude Code \cite{anthropic2025claudecode}, Codex CLI \cite{openai2025codex}, and OpenClaw \cite{openclaw2026} fundamentally depends on the underlying model's ability to solve diverse terminal tasks.

\begin{figure}[t]
    \centering
    \includegraphics[width=0.9\columnwidth]{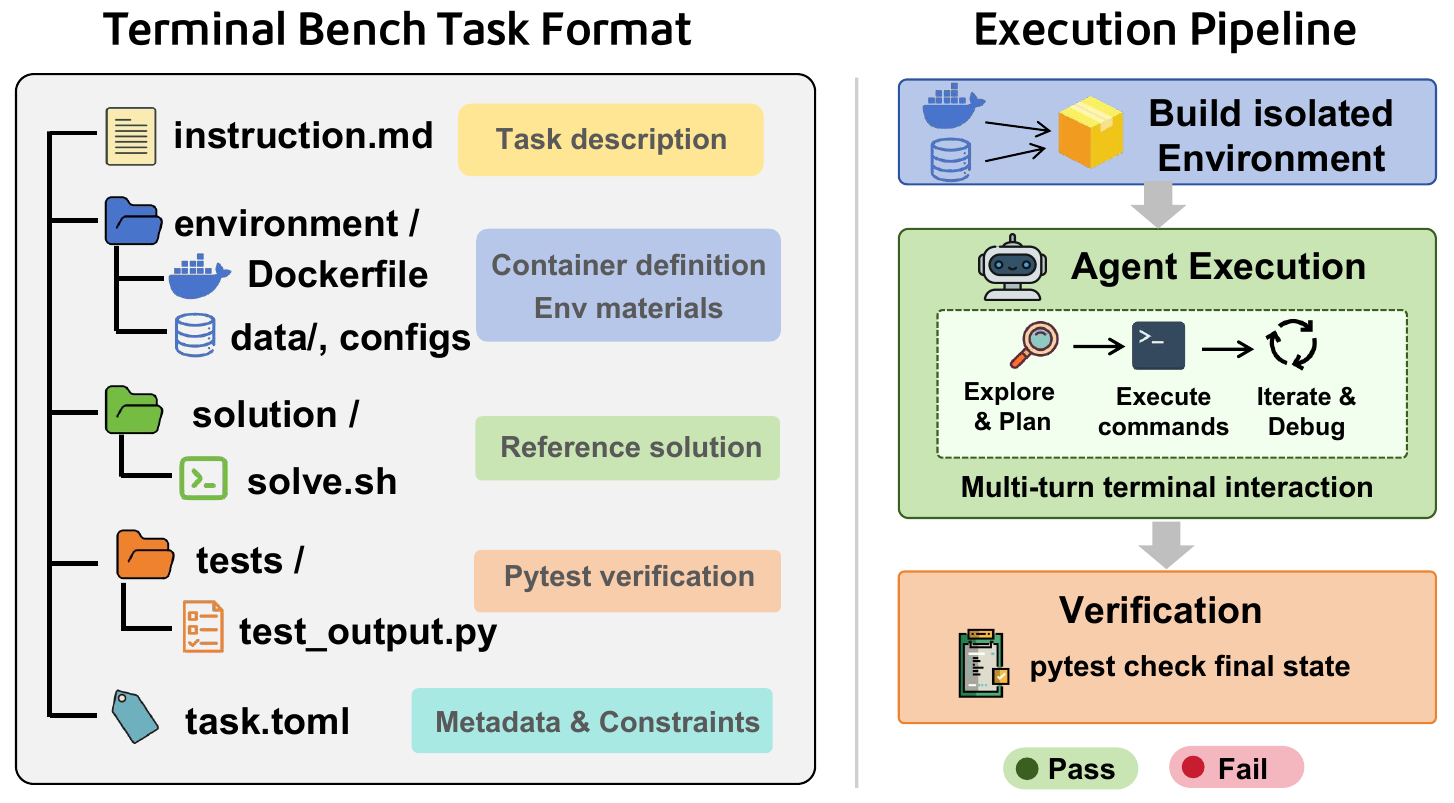}
    \caption{Left: the file structure of a Terminal-Bench task package. Right: the evaluation workflow consisting of environment building, agent execution, and automated verification.}
    \label{fig:tb-format}
\end{figure}

Terminal-Bench \cite{merrill2026terminal} provides a standard benchmark for evaluating such capabilities, spanning domains from software engineering to scientific computing. As shown in Figure~\ref{fig:tb-format}, each task is defined by a tightly integrated package including a task description, an environment definition, a reference solution, and verification tests. During evaluation, the agent iteratively interacts with the terminal inside an isolated container built from the task-defined environment to complete the task. Succeeding on this benchmark requires more than surface-level command generation: the agent must interpret complex outputs, diagnose failures, and adaptively revise its strategy \cite{gandhi2026endless,pi2026data}. Yet even advanced proprietary models achieve only moderate scores, underscoring the fundamental difficulty of the challenge~\cite{wu2026large,pi2026data}.

Training such agentic terminal capabilities demands high-quality, verifiable, and diverse terminal tasks and interaction trajectories \cite{fan2026toward}. Recent research on terminal task synthesis has primarily proceeded along two lines. The first mines task instances from real code repositories \cite{wu2026large,lin2026cli}. While producing executable environments, the resulting tasks remain largely confined to SWE-style~\cite{jimenez2024swe,wang2025swe,yang2026swe} bug fixes and issue resolution, exhibiting limited \textit{diversity and scalability}. Another line generates tasks through LLM-driven synthesis pipelines \cite{zhu2026termigen,pi2026data,fan2026toward}. These methods typically synthesize different task components in separate stages without real-time terminal feedback, making it difficult to guarantee internal consistency and \textit{execution verifiability} of the generated tasks. These limitations raise a central question: can we leverage the inherent agentic capabilities of LLMs to scalably generate terminal tasks that cover diverse domains and expression forms while maintaining task reliability?

We observe a key fact: \textbf{\textit{the task synthesis process itself, which involves gathering or processing relevant materials, designing problems, writing code and environment configurations, constructing verification tests, and iteratively checking consistency, is inherently a complex agentic terminal task.}} This observation inspires us to propose \textbf{Meta-Task}: redefining terminal task synthesis as a Terminal-Bench-format terminal task (which we call a \textit{meta-task}), completed end-to-end by an agent inside a Docker container. Specifically, Meta-Task encapsulates each synthesis request as a standard Terminal-Bench-format task, containing an instruction that defines the target task requirements and guides the agent through the creation workflow, a pre-configured Docker environment (with example tasks, optional external material support, and output templates), and tests that verify the generated output. Within this environment, the agent autonomously explores examples, gathers materials, designs and implements a complete task package (instruction, Dockerfile, solution, tests), and iteratively refines the output through a self-validation loop until all checks pass.

This design yields three core advantages. (1) \textbf{Inherent execution grounding}: the generation process is itself grounded in a real terminal environment, where the agent contends with actual dependency installations, file operations, and test failures, substantially reducing the hallucination issues of pure LLM generation. (2) \textbf{High scalability}: the framework directly leverages the powerful agentic capabilities of LLMs for task synthesis, with a flexible architecture that supports diverse input materials and enables efficient batch generation and easy extension to new domains. (3) \textbf{Natural compatibility with advanced agent scaffolds}: since meta-tasks follow the standard Terminal-Bench format, the framework can directly leverage state-of-the-art agent scaffolds (e.g., Claude Code \cite{anthropic2025claudecode}, Codex CLI \cite{openai2025codex}) and benefit from their sophisticated context management, tool orchestration, and long-horizon planning capabilities \cite{bui2026building,harness_survey}.

To further enhance the diversity, realism, and quality of synthesized data, we develop three complementary components. First, we decouple target task requirements along multiple dimensions and introduce a \textbf{multi-phase synthesis mode} that dynamically designs novel task specifications, substantially expanding synthesis coverage. Second, we provide \textbf{optional external material support}, including a configurable search-and-download tool and support for pre-supplied resources, enabling the agent to ground synthesized tasks in real-world code and data. Third, we apply \textbf{LLM-as-Judge filtering} on sampled trajectories to ensure the quality of the final training data. In summary, our main contributions are as follows:
\begin{itemize}[nosep, leftmargin=*]
    \item We propose Meta-Task, a framework that redefines terminal task synthesis as a Terminal-Bench-format terminal task. By having an agent complete task creation end-to-end inside a real container environment, we achieve inherent execution grounding and high scalability.
    \item We develop a decoupled diversity control mechanism with multi-phase synthesis, optional external material support, and LLM-as-Judge trajectory filtering, jointly enhancing task diversity, realism, and training data quality.
    \item Fine-tuning on only 3,221 Meta-Task synthesized trajectories achieves \textbf{22.5\%} and \textbf{31.8\%} Avg Pass@1 for Qwen3-14B and Qwen3-32B respectively on Terminal-Bench 2.0, outperforming concurrent approaches with significantly less training data.
\end{itemize}

%% file: sections/method.tex
\section{Meta-Task Pipeline}
\label{sec:method}

\begin{figure*}[t]
  \centering
  \includegraphics[width=0.90\textwidth]{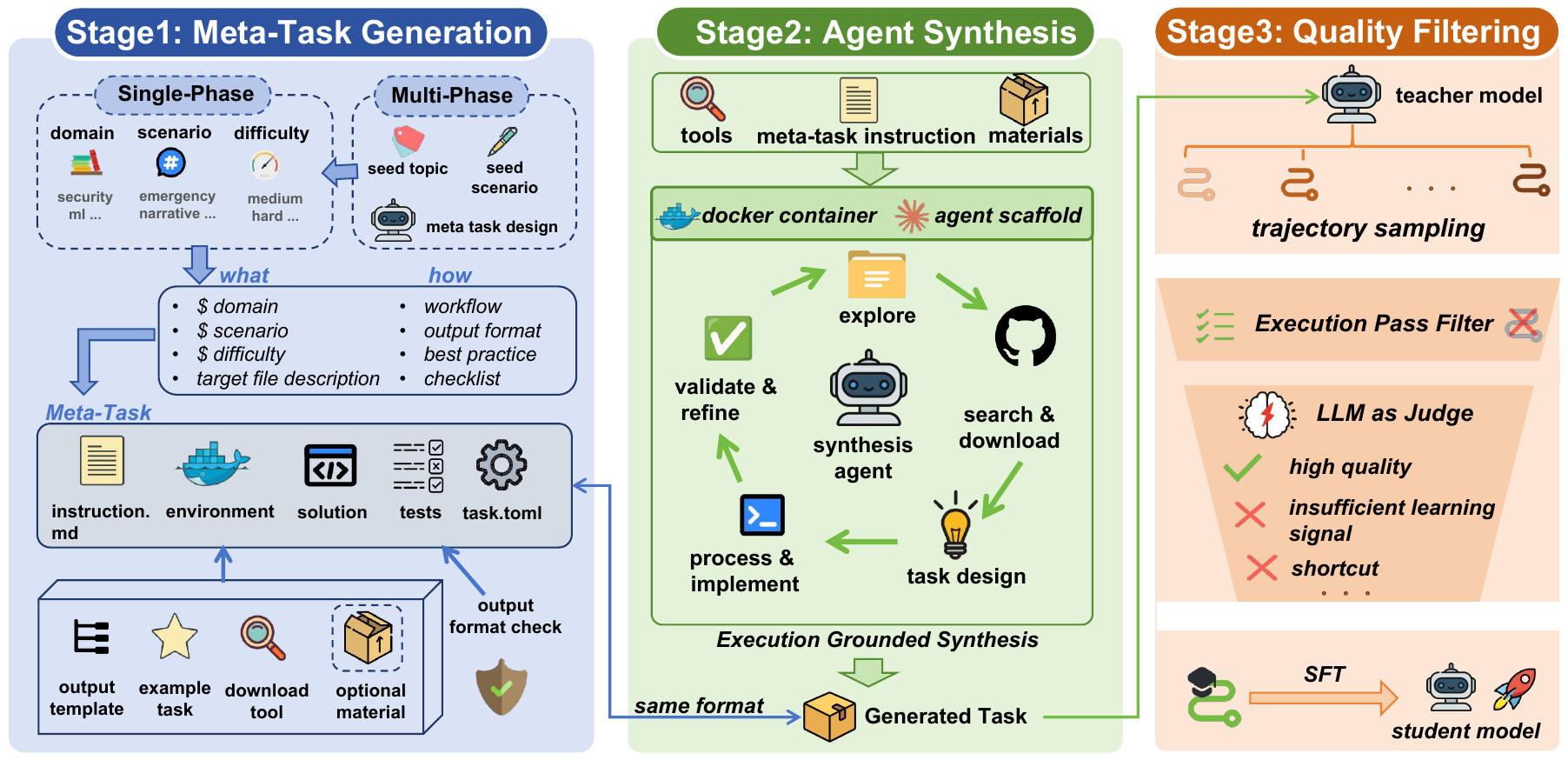}
  \caption{Overview of the Meta-Task pipeline. \textbf{Stage 1 (Meta-Task Generation)}: decoupled diversity dimensions (category, scenario, difficulty) are composed into a meta-task instruction via single-phase composition or multi-phase dynamic task specification design, paired with a pre-configured Docker environment. \textbf{Stage 2 (Agent Synthesis)}: an agent executes the meta-task inside a Docker container, exploring examples, searching real materials, designing and implementing new tasks, and iteratively self-validating until all checks pass. \textbf{Stage 3 (Quality Filtering)}: a teacher model samples trajectories on generated tasks; passed trajectories undergo LLM-as-Judge screening to produce the final SFT training data.}
  \label{fig:pipeline}
\end{figure*}

\subsection{Overview}
\label{sec:overview}

As illustrated in Figure~\ref{fig:pipeline}, the Meta-Task pipeline consists of three stages. \textbf{Stage 1: Meta-Task Generation} (\S\ref{sec:generation}): each meta-task is constructed as a Terminal-Bench-format task package with a synthesis instruction, a Docker environment, and verification tests; diversity is achieved through decoupled dimension templates and a multi-phase mechanism that dynamically designs novel task specifications. \textbf{Stage 2: Agent Synthesis} (\S\ref{sec:execution}): an agent executes the meta-task inside a Docker container, autonomously exploring, searching, designing, implementing, and iteratively self-validating until all checks pass. \textbf{Stage 3: Quality Filtering} (\S\ref{sec:quality}): a teacher model samples trajectories on generated tasks; passed trajectories undergo LLM-as-Judge screening to produce the final SFT training data.

\subsection{Meta-Task Generation}
\label{sec:generation}

\subsubsection{Synthesis Instruction}

The meta-task instruction is constructed by combining fixed and variable components (the full template is provided in Appendix~\ref{app:prompts}). The \textbf{fixed components} define \textbf{\textit{how}} to synthesize, specifying target file descriptions, the synthesis workflow, output format requirements, best practices (e.g., Dockerfile conventions, test design guidelines), and a self-validation checklist. The \textbf{variable components} control \textbf{\textit{what}} to synthesize: three independently selectable dimension templates covering the technical domain and task pattern ideas (category), instruction writing style (scenario), and complexity requirements (difficulty). At each generation, these variable parts are randomly selected and combined, with internal lists shuffled to prevent ordering bias.

\subsubsection{Diversity Control}
\label{sec:diversity}

\paragraph{Category (Technical Domain).} We predefine 39 category templates covering a wide range of domains from low-level systems programming and network administration to high-level ML pipelines and cross-disciplinary tasks. Each template provides domain-specific guidance including representative task patterns, recommended tools, and verification approaches. The sampling probability of each category is configurable, allowing flexible control over the domain distribution of generated tasks.

\paragraph{Scenario (Instruction Writing Style).} We design 10 scenario templates simulating diverse real-user communication patterns, from ultra-minimal one-liner goals (\textit{minimal}) to structured requirements (\textit{structured\_request}), workplace narratives (\textit{narrative}), and urgent on-call notes (\textit{emergency}). This dimension captures the stylistic diversity of real-world terminal task descriptions.

\paragraph{Difficulty Level.} Four difficulty levels (\textit{easy, medium, hard, extreme}) are defined, each specifying expected knowledge depth, solution complexity, and an anti-pattern list of task types to avoid.

These three dimensions are fully orthogonal, yielding $39 \times 10 \times 4 = 1{,}560$ distinct base combinations in single-phase mode (examples in Appendix~\ref{app:categories}).

\paragraph{Multi-Phase Synthesis.}
\label{sec:multiphase}
To transcend the predefined template space, we introduce a multi-phase mode. First, a lightweight meta-task instructs an agent to design a new (category, scenario) specification pair based on two seed signals: a \textit{topic seed} sampled from approximately 2,000 fine-grained technical topics, and a \textit{scenario constraint} sampled from approximately 120 style descriptions. Then, the generated specifications are injected into the standard meta-task instruction, replacing the variable parts while keeping all fixed parts unchanged. This extends the effective diversity space to over 240,000 seed combinations, and since the agent can creatively interpret each seed, the achievable diversity is theoretically unbounded and trivially extensible by adding new topics.

\begin{table*}[t]
  \centering
  \small
  \resizebox{\textwidth}{!}{%
  \begin{tabular}{@{}lcccccc@{}}
  \toprule
  \textbf{Property} & \textbf{TerminalTraj} & \textbf{TermiGen} & \textbf{Nemotron-Terminal} & \textbf{CLI-Gym} & \textbf{SkillSynth} & \textbf{Meta-Task} \\
  & \cite{wu2026large} & \cite{zhu2026termigen} & \cite{pi2026data} & \cite{lin2026cli} & \cite{fan2026toward} & (Ours) \\
  \midrule
  Real-world material & Required & $\times$ & Optional & Required & Required & Optional \\
  Diversity control & $\times$ & $\checkmark$ & $\checkmark$ & $\times$ & $\circ$ & $\checkmark$ \\
  Execution grounded & $\circ$ & $\circ$ & $\times$ & $\circ$ & $\circ$ & $\checkmark$ \\
  Agent scaffold compatible & $\times$ & $\times$ & $\times$ & $\times$ & $\times$ & $\checkmark$ \\
  Quality filtering & $\checkmark$ & $\circ$ & $\circ$ & $\checkmark$ & $\circ$ & $\checkmark$ \\
  \midrule
  \# Synthesized tasks & 32.3k & 3.5k & 264.2k & 1.6k & 3.5k & 15.0k \\
  \# SFT trajectories & 50.7k & 3.3k & 490.5k & 0.3k & 10.6k & 3.2k \\
  \bottomrule
  \end{tabular}%
  }
  \caption{Comparison of terminal task synthesis approaches. $\checkmark$ = yes, $\times$ = no, $\circ$ = partial.}
  \label{tab:comparison}
  \end{table*}

\subsubsection{Environment Configuration}

Beyond the instruction, the Docker environment for each meta-task is pre-configured with:

\begin{itemize}[nosep,leftmargin=*]
    \item \textbf{Task skeleton template}: a complete directory structure with all required files pre-created as placeholders, transforming generation into a structured content-filling task.
    \item \textbf{Example tasks}: one task randomly sampled from a pool of 100 high-quality reference tasks (synthesized by Claude Opus 4.6 and manually verified), serving as a concrete format reference.
    \item \textbf{Search-and-download tool} (optional): a lightweight tool enabling the agent to retrieve real open-source materials from GitHub and HuggingFace during generation, grounding tasks in authentic development scenarios.
    \item \textbf{Pre-supplied materials} (optional): domain-specific input data or resources provided directly in the environment when available.
\end{itemize}

The remaining components are fixed: a base Dockerfile with common development tools and appropriate execution time limits.

\subsection{Agent Synthesis}
\label{sec:execution}

Once a meta-task is assembled, an agent is launched inside the built Docker container with a compatible scaffold (e.g., Claude Code \cite{anthropic2025claudecode}). The agent will \textbf{explore} the provided example and template to understand the expected format, \textbf{search} for relevant real-world materials using the built-in tool, \textbf{design} a problem based on the gathered materials, and \textbf{implement} a complete task package (instruction, Dockerfile, solution, tests). The agent is also required to validate its output by executing the solution, running tests, and checking cross-component consistency, ensuring that the generated task is internally coherent, executable, and free of answer leakage. Since this entire process takes place within a real terminal environment with authentic stdout/stderr feedback, it fundamentally differs from approaches where generation and execution are separated or absent.

\subsection{Quality Filtering}
\label{sec:quality}

We instantiate the test script as an automated task completeness checker, which verifies the structural integrity and internal consistency of each synthesized task package, serving as a first-pass quality filter. We then use the teacher model to sample solution trajectories on the validated tasks and apply further quality filtering to obtain reliable training data.

\paragraph{Execution-based filtering.} Only trajectories where the agent's solution passes all verification tests are retained.

\paragraph{LLM-as-Judge filtering.} The passed trajectories undergo further screening by an LLM reviewer, which identifies and removes trajectories where suggestive code annotations or solution leakage reduce task difficulty, or where trivially short interactions lack sufficient learning signal. The trajectories surviving both filtering layers constitute the final SFT training data (prompt in Appendix~\ref{app:prompts}).

\subsection{Implementation and Data Statistics}
\label{sec:statistics}

\subsubsection{Implementation Details}

We use the open-source Qwen3.5-397B-A17B-FP8 as the backbone model throughout the pipeline. For task synthesis (Stage 2), we deploy it locally via vLLM \cite{kwon2023efficient} and integrate it with the Claude Code \cite{anthropic2025claudecode} agent scaffold, which provides advanced context management and tool orchestration. We adapt the Harbor \cite{Harbor_Framework} scheduling framework to manage meta-task execution and output collection. For trajectory sampling, we use the Terminus-2 agent framework. In both stages, the model operates in non-thinking mode with temperature 0.7 following official recommendations. For quality filtering, we adopt trajectory-level LLM-as-Judge review using the same model in thinking mode. We choose trajectory-level over task-level filtering because evaluating isolated code fragments without full execution context makes it difficult for the judge model to reliably assess task quality, leading to false positives and missed issues. Trajectory-level filtering is more precise, targeting only trajectories where the agent actually exhibited undesirable behavior.

\subsubsection{Data Statistics}

Using our framework, we synthesize approximately 15,000 complete task packages in total, with a synthesis success rate above 85\% on average across difficulty settings, indicating that the synthesis stage itself is reliable. We then sample 14,040 trajectories on these tasks using the teacher model, of which 5,004 pass all verification tests (pass rate: 35.6\%). Note that this pass rate measures how often the teacher solves our tasks rather than how often synthesis succeeds, so its low value indirectly reflects the non-trivial difficulty of the synthesized tasks. After strict LLM-as-Judge quality filtering, 3,221 trajectories are retained as the final training data.

We further run a decontamination check between our synthesized tasks and the Terminal-Bench test tasks, comparing every instruction pair by word-level n-gram overlap ($n=3,5,8$), TF-IDF cosine similarity, and technical-entity matching. No pair is flagged as high risk: n-gram scores are essentially zero, and the highest-scoring pairs share only generic domain terms while differing entirely in goals, inputs, and required solutions, which we verified by manual inspection.

\subsubsection{Comparison with Existing Approaches}

Table~\ref{tab:comparison} summarizes the key design differences between Meta-Task and existing terminal task synthesis approaches. Compared to repository-centric approaches \cite{wu2026large,lin2026cli} that are constrained by existing codebases, Meta-Task can synthesize tasks independently while optionally grounding them in real materials via a search tool. Unlike staged pipelines \cite{zhu2026termigen,pi2026data,fan2026toward} where task generation and execution verification are decoupled, our approach performs the entire synthesis within a live Docker container with iterative self-verification, ensuring internal consistency and eliminating hallucination issues inherent in pure generation without execution grounding. The decoupled diversity control mechanism further provides explicit, composable control over task characteristics. Additionally, the standard Terminal-Bench format design ensures high scalability and natural compatibility with advanced agent scaffolds.

%% file: sections/experiment.tex
\section{Experiments}
\label{sec:experiments}

\subsection{Experimental Setup}

\paragraph{Training and Evaluation.} We use Qwen3-14B and Qwen3-32B \cite{yang2025qwen3} as our base models and perform supervised fine-tuning via LoRA \cite{hu2022lora} using the SWIFT framework \cite{zhao2025swift}. We evaluate on Terminal-Bench 2.0 \cite{merrill2026terminal} following the official evaluation protocol, employing Harbor \cite{Harbor_Framework} as the orchestration framework together with Terminus-2 as the agent scaffold. All reported results are the mean over three independent runs. Full training and inference details are provided in Appendix~\ref{app:training}.

\paragraph{Baselines.} We compare against three categories of systems. \textbf{Closed-source models}: we include leading proprietary systems spanning GPT-5.3-Codex, Claude Opus 4.6, Gemini 3 Pro, and other frontier models from the official Terminal-Bench leaderboard. \textbf{Open-source models}: we report results for a range of open-source models including Qwen3.5-397B-A17B (our teacher model), GLM-5, DeepSeek-V3.2, and the Qwen3-14B/32B models that serve as our fine-tuning backbones. \textbf{Fine-tuned models}: we compare with recent concurrent works on terminal task synthesis, including TerminalTraj \cite{wu2026large}, CLI-Gym \cite{lin2026cli}, TermiGen \cite{zhu2026termigen}, Nemotron-Terminal \cite{pi2026data} and SkillSynth \cite{fan2026toward}.

\subsection{Main Results}

\begin{table*}[t]
\centering
\small
\setlength{\tabcolsep}{12pt}
\begin{tabular}{@{}l|lccc}
\toprule
& \textbf{Model} & \textbf{Size} & \textbf{Scaffold} & \textbf{Avg Pass (\%)} \\
\midrule
\multirow{6}{*}{\textbf{\textit{Closed-Source}}}
& GPT-5.3-Codex & -- & Terminus-2 & 64.7$^\dagger$ \\
& Claude Opus 4.6 & -- & Terminus-2 & 62.9$^\dagger$ \\
& Gemini 3 Pro & -- & Terminus-2 & 56.9$^\dagger$ \\
& Claude Sonnet 4.5 & -- & Terminus-2 & 42.8$^\dagger$ \\
& GPT-5 & -- & Terminus-2 & 35.2$^\dagger$ \\
& Gemini 2.5 Pro & -- & Terminus-2 & 32.6$^\dagger$ \\
\midrule
\multirow{9}{*}{\textbf{\textit{Open-Source}}}
& GLM-5 & 744B & Terminus-2 & 52.4$^\dagger$ \\
& Kimi K2.5 & 1T & Terminus-2 & 43.2$^\dagger$ \\
& Minimax M2.5 & 229B & Terminus-2 & 42.2$^\dagger$ \\
& DeepSeek-V3.2 & 685B & Terminus-2 & 39.6$^\dagger$ \\
& GPT-OSS & 120B & Terminus-2 & 18.7$^\dagger$ \\
& Qwen3-Coder-480B-A35B & 480B & Terminus-2 & 23.9$^\dagger$ \\
& Qwen3.5-397B-A17B & 397B & Terminus-2 & 52.5$^*$ \\
& Qwen3-14B \textbf{\textit{(base)}} & 14B & Terminus-2 & 5.2 \\
& Qwen3-32B \textbf{\textit{(base)}} & 32B & Terminus-2 & 4.1 \\
\midrule
\multirow{9}{*}{\textbf{\textit{Fine-tuned}}}
& TermiGen-32B \cite{zhu2026termigen} & 32B & BashAgent & 19.3$^*$ \\
& LiberCoder-32B \cite{lin2026cli} & 32B & OpenHands & 19.5$^*$ \\
& LiberCoder-235B-A22B \cite{lin2026cli} & 235B & OpenHands & 31.0$^*$ \\
& TerminalTraj-14B \cite{wu2026large} & 14B & Terminus-2 & 19.1$^*$ \\
& TerminalTraj-32B \cite{wu2026large} & 32B & Terminus-2 & 22.0$^*$ \\
& Nemotron-Terminal-14B \cite{pi2026data} & 14B & Terminus-2 & 20.2$^*$ \\
& Nemotron-Terminal-32B \cite{pi2026data} & 32B & Terminus-2 & 27.4$^*$ \\
& SkillSynth-14B \cite{fan2026toward} & 14B & Terminus-2 & 19.9$^*$ \\
& SkillSynth-32B \cite{fan2026toward} & 32B & Terminus-2 & 29.6$^*$ \\
\midrule
\multirow{2}{*}{\textit{\textbf{Ours}}}
& \cellcolor{gray!15}\textbf{Meta-Task-14B} & \cellcolor{gray!12}14B & \cellcolor{gray!12}Terminus-2 & \cellcolor{gray!12}\textbf{22.5} \\
& \cellcolor{gray!15}\textbf{Meta-Task-32B} & \cellcolor{gray!12}32B & \cellcolor{gray!12}Terminus-2 & \cellcolor{gray!12}\textbf{31.8} \\
\bottomrule
\end{tabular}
\caption{Results on Terminal-Bench 2.0. $^\dagger$ denotes results from the official leaderboard and $^*$ denotes results cited from the respective original reports.}
\label{tab:main_results}
\end{table*}

Table~\ref{tab:main_results} presents the main results on Terminal-Bench 2.0. Fine-tuning on Meta-Task-synthesized data brings significant improvements over the base models: Qwen3-14B improves from 5.2\% to 22.5\% (+17.3) and Qwen3-32B from 4.1\% to 31.8\% (+27.7), validating the effectiveness of our synthesized training data. Notably, Meta-Task-32B even surpasses Qwen3-Coder-480B-A35B (23.9\%), a model with over 15$\times$ more parameters, demonstrating that targeted terminal task training can compensate for a substantial gap in model scale.

Compared with concurrent data synthesis and fine-tuning approaches, Meta-Task-32B (31.8\%) outperforms Nemotron-Terminal-32B (27.4\%) and TerminalTraj-32B (22.0\%) at the same model scale, and closely approaches LiberCoder-235B (31.0\%), which uses a model with over 7$\times$ more total parameters. For the 14B scale, Meta-Task-14B (22.5\%) outperforms Nemotron-Terminal-14B (20.2\%) and TerminalTraj-14B (19.1\%). These results are achieved with only 3,221 training trajectories, significantly fewer than TerminalTraj (50.7K) and Nemotron-Terminal (490.5K), demonstrating the data efficiency of our approach.

While frontier closed-source systems such as GPT-5.3-Codex (64.7\%) and Claude Opus 4.6 (62.9\%) still hold a significant lead, Meta-Task-32B (31.8\%) approaches GPT-5 (35.2\%) and Gemini 2.5 Pro (32.6\%), substantially narrowing the gap between small open-source fine-tuned models and full-scale proprietary systems.

\begin{figure}[t]
  \centering
  \includegraphics[width=0.9\columnwidth]{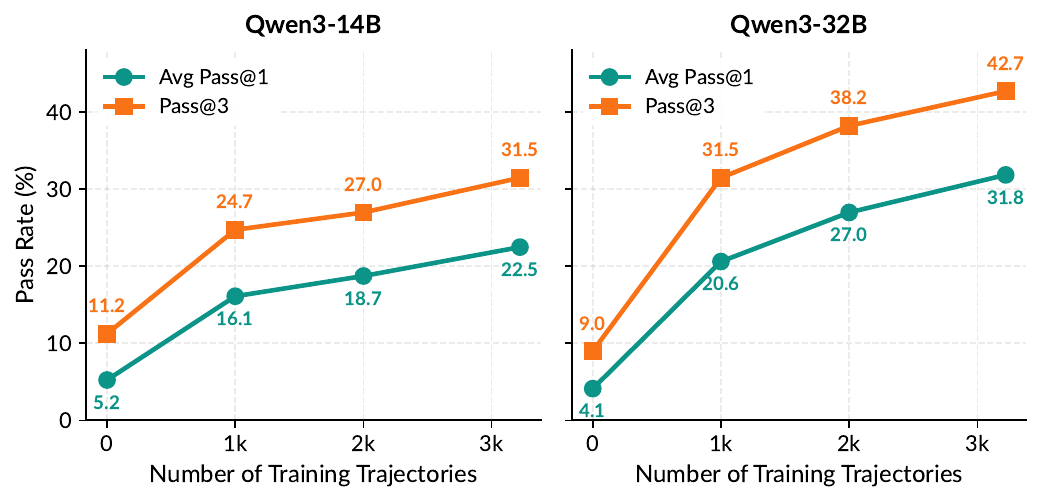}
  \caption{Scaling behavior of Meta-Task-synthesized data. Avg Pass@1 and Pass@3 are plotted against the number of training trajectories for both Qwen3-14B (left) and Qwen3-32B (right).}
  \label{fig:scaling}
\end{figure}

\begin{figure*}[t]
  \centering
  \includegraphics[width=\textwidth]{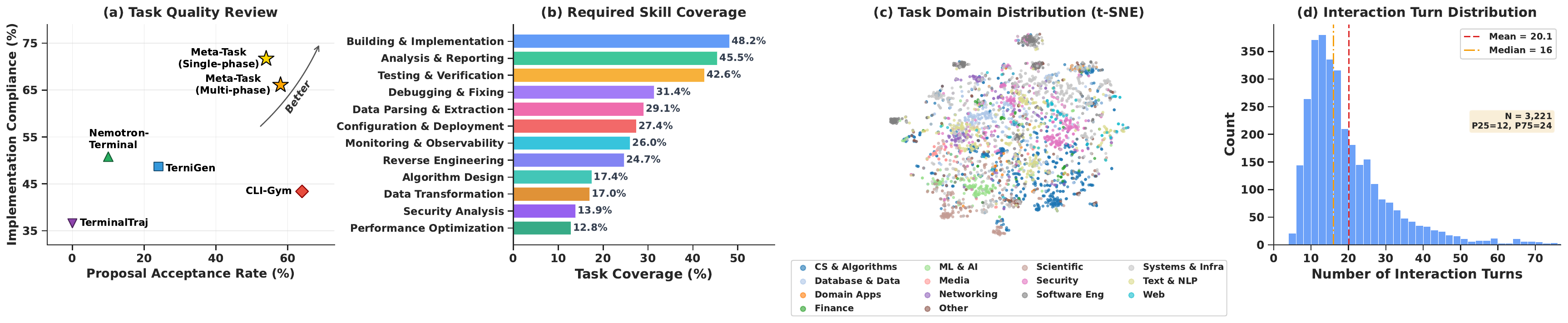}
  \caption{Task quality, diversity, and trajectory analysis. (a) Task quality evaluation across synthesis methods (50 sampled tasks per method, scored by Claude Opus 4.6 on the Terminal-Bench Rubric). (b) Skill distribution across 3,221 tasks (multi-label). (c) t-SNE of task instructions by domain. (d) Distribution of interaction turns.}
  \label{fig:task_analysis}
\end{figure*}

\subsection{Scaling Behavior}
\label{sec:scaling}

We investigate the scaling behavior of Meta-Task-synthesized data along three dimensions: data volume, model size, and test-time compute (Pass@$k$). We train both Qwen3-14B and Qwen3-32B on progressively increasing subsets of the synthesized trajectories (0, 1000, 2000, 3221) and report Avg Pass@1 and Pass@3 at each scale.

As shown in Figure~\ref{fig:scaling}, three scaling trends emerge: (1) \textbf{Data scaling}: both models show consistent performance gains as training data volume increases, indicating that more synthesized trajectories continue to benefit model performance. (2) \textbf{Model scaling}: the 32B model benefits more from the same data, achieving higher absolute performance and steeper improvement curves than 14B. (3) \textbf{Test-time scaling}: Pass@3 consistently outperforms Pass@1, demonstrating that the fine-tuned models develop diverse problem-solving strategies benefiting from multiple attempts.

\begin{table}[t]
\centering
\small
\begin{tabular}{@{}lccc@{}}
\toprule
\textbf{Configuration} & \textbf{\# Trajs} & \textbf{Avg Pass@1} & \textbf{Pass@3} \\
\midrule
Passed only & 5,004 & 28.5 & 39.3 \\
Passed + Judge & 3,221 & \textbf{31.8} & \textbf{42.7} \\
\bottomrule
\end{tabular}
\caption{Effect of trajectory-level quality filtering (Qwen3-32B, Terminal-Bench 2.0).}
\label{tab:ablation_filtering}
\end{table}

\subsection{Ablation: Quality over Quantity}
\label{sec:ablation_filtering}

We conduct an ablation study to validate the effectiveness of our quality filtering component on Qwen3-32B, comparing 5,004 trajectories that only pass execution tests against 3,221 trajectories that additionally undergo strict LLM-as-Judge quality filtering.

As shown in Table~\ref{tab:ablation_filtering}, applying LLM-as-Judge filtering yields +3.3 Avg Pass@1 and +3.4 Pass@3 gains despite using fewer trajectories. The filtered trajectories primarily include cases where suggestive code annotations or overly complete expected-output files reduce task difficulty, causing the agent to take shortcuts, as well as trajectories with short interaction lengths that lack sufficient learning signal. The LLM-as-Judge effectively identifies and removes these successful but low-quality trajectories. This result confirms that trajectory quality matters more than quantity for training terminal agents, consistent with findings in CLI-Gym \cite{lin2026cli}.

\begin{table*}[t]
\centering
\small
\renewcommand{\arraystretch}{1.2}
\begin{tabular}{@{}m{3.8cm}rm{9.5cm}@{}}
\toprule
\textbf{Error Category} & \textbf{Ratio} & \textbf{Description} \\
\midrule
Context Compaction Loss & 47.3\% & Loses task constraints after context summarization, submitting incorrect solutions. \\
Flawed Self-Verification & 32.9\% & Inline checks pass but miss edge cases caught by the test suite. \\
Step Budget Exhaustion & 15.1\% & Explores multiple strategies without converging before the turn limit. \\
Repetitive Strategy Loop & 4.8\% & Repeats the same failing approach with minor variations. \\
\bottomrule
\end{tabular}
\renewcommand{\arraystretch}{1.0}
\caption{Error analysis of failed trajectories (model-attributable only).}
\label{tab:error-analysis}
\end{table*}

\subsection{Task and Trajectory Analysis}
\label{sec:task_analysis}

We further analyze the quality of our synthesized tasks, their diversity, and the characteristics of the SFT trajectories (a concrete task example is provided in Appendix~\ref{app:task_examples}). Figure~\ref{fig:task_analysis} presents the results across four complementary dimensions.

\paragraph{Task quality comparison.} We assess synthesized task quality using the official Terminal-Bench review pipeline with Claude Opus 4.6 as the automated reviewer, which includes a \textit{Proposal review} evaluating task ideas and an \textit{Implementation review} scoring the full task package against 19 criteria (e.g., verifiability, difficulty, novelty, anti-cheat robustness). For each method, we randomly sample 50 tasks. As shown in Figure~\ref{fig:task_analysis}(a), Meta-Task achieves the highest implementation compliance (66--72\%) with strong proposal acceptance (54--58\%). CLI-Gym scores higher on proposal acceptance (64\%) but only 43\% on implementation due to lacking solution scripts. Other methods score lower on both axes. The evaluation protocol is described in Appendix~\ref{app:task_quality}.

\paragraph{Skill coverage.} As shown in Figure~\ref{fig:task_analysis}(b), we categorize required skills into 12 high-level competency areas (multi-label). The distribution reveals that our synthesized tasks go far beyond the narrow bug-fixing pattern prevalent in repository-mined datasets: the skills are evenly spread across all categories, with specialized competencies such as Reverse Engineering (24.7\%) and Security Analysis (13.9\%) maintaining substantial representation alongside dominant ones like Building \& Implementation (48.2\%).

\paragraph{Domain diversity.} Figure~\ref{fig:task_analysis}(c) visualizes the task instruction embeddings via t-SNE, colored by 13 top-level technical domains. The embedding space exhibits broad and uniform coverage with well-separated domain clusters. Notably, several domains manifest multiple distinct sub-clusters, indicating that our synthesis mechanisms produce diverse instantiations within the same domain rather than repetitive variations.

\paragraph{Trajectory characteristics.} Figure~\ref{fig:task_analysis}(d) shows the distribution of interaction turns across 3,221 filtered trajectories (mean=20, median=16). The majority of trajectories complete in 6--20 turns, reflecting moderate-complexity tasks requiring planning, execution, debugging, and verification. A substantial long tail extends beyond 25 turns, providing rich learning signal for long-horizon problem-solving. The absence of trivially short trajectories (fewer than 5 turns) confirms that our quality filtering effectively removes low-signal examples.

\subsection{Error Analysis}

To understand the remaining limitations, we analyze 146 model-attributable failed trajectories across 3 independent evaluation runs, excluding infrastructure errors and wall-clock timeouts (detailed analysis in Appendix~\ref{app:error}). As shown in Table~\ref{tab:error-analysis}, the dominant failure mode is Context Compaction Loss (47.3\%): as the trajectory grows, earlier observations are summarized away, causing the agent to lose critical task constraints. The second most common issue is Flawed Self-Verification (32.9\%), where ad-hoc checks pass but miss edge cases caught by the actual test suite. Notably, 80.1\% of all failures involve the agent marking the task as complete despite an incorrect solution, indicating that miscalibrated self-assessment is the primary weakness to address in future work.

%% file: sections/related_work.tex
\section{Related Work}

\paragraph{Terminal Agents.} LLM-based agents that operate through command-line interfaces have become a central paradigm for real-world task automation \cite{yao2023react,yang2024sweagent,wang2025openhands,anthropic2025claudecode,openai2025codex}. To systematically assess these capabilities, a series of SWE-like benchmarks \cite{jimenez2024swe,wang2025swe,yang2026swe,zan2025multi,chen2026sweuniv} and terminal interaction benchmarks \cite{lin2018nl2bash,yang2023intercode,merrill2026terminal} have been proposed to evaluate agents across diverse task domains. The practical effectiveness of these systems fundamentally depends on the underlying model's terminal capabilities \cite{cheng2026llm,ren2026self,xia2025live}, and training such capabilities requires high-quality, verifiable, and diverse terminal task data.

\paragraph{Terminal Task Synthesis.} Existing approaches follow two paradigms. The first extracts tasks from real-world codebases: TerminalTraj \cite{wu2026large} scores and filters GitHub repositories at scale, and CLI-Gym \cite{lin2026cli} inverts healthy environments into faulty states. The second uses LLMs to synthesize tasks directly \cite{qin2023toolllm}: TermiGen \cite{zhu2026termigen} employs a multi-agent framework for staged generation, Nemotron-Terminal \cite{pi2026data} combines dataset adaptation with skill-oriented synthesis, and SkillSynth \cite{fan2026toward} controls diversity through skill graph path sampling. Additionally, SETA \cite{seta}, Endless Terminals \cite{gandhi2026endless}, and OpenClaw-rl \cite{wang2026openclaw} explore reinforcement learning based approaches, while Agent-World \cite{dong2026agent} and Meta-Harness \cite{lee2026meta} address environment synthesis and harness optimization respectively.

%% file: sections/conclusion.tex
\section{Conclusion}

We proposed Meta-Task, a framework that redefines terminal task synthesis as a Terminal-Bench-format terminal task itself, where an agent operates end-to-end inside a Docker container to generate and self-validate complete task packages, achieving inherent execution grounding and high scalability that address the key limitations of prior approaches. Together with decoupled diversity control, multi-phase synthesis, external material support, and LLM-as-Judge trajectory filtering, Meta-Task jointly enhances task diversity, realism, and training data quality. Fine-tuning on only 3,221 synthesized trajectories yields strong performance on Terminal-Bench 2.0, outperforming concurrent approaches with significantly less data.

%% file: sections/limitation.tex
\section*{Limitations}

Our work focuses exclusively on Linux-based terminal task synthesis within Docker containers. We have not yet extended the framework to other operating system ecosystems such as Windows or macOS, where different toolchains, shell environments, and system APIs would require adapted environment configurations and verification approaches. Additionally, while we demonstrate strong diversity through template-based control and multi-phase synthesis, the seed sources for diversity (topic lists, scenario constraints) are currently curated manually. Incorporating broader seed sources from real-world developer activity, such as Stack Overflow questions, GitHub issue discussions, or technical documentation, could further expand the synthesis coverage.

%% file: sections/appendix.tex
\newpage
\appendix

\section{Training and Inference Details}
\label{app:training}
\paragraph{Fine-tuning.} Table~\ref{tab:hyperparams} lists the hyperparameters for LoRA fine-tuning. Both Qwen3-14B and Qwen3-32B are fine-tuned in non-thinking mode using the SWIFT framework \cite{zhao2025swift} on 8$\times$ A100 80GB GPUs with DeepSpeed ZeRO-3.

\begin{table}[h]
\centering
\small
\begin{tabular}{@{}ll@{}}
\toprule
\textbf{Hyperparameter} & \textbf{Value} \\
\midrule
LoRA rank / alpha & 64 / 128 \\
LoRA dropout & 0.05 \\
LoRA target modules & all linear layers \\
Learning rate & 1e-4 \\
LR scheduler & cosine (warmup 5\%) \\
Sequence length & 40,960 \\
Epochs & 5 \\
Batch size (per device) & 1 \\
Gradient accumulation & 1 \\
Optimizer & AdamW ($\beta_1$=0.9, $\beta_2$=0.95) \\
Weight decay & 0.1 \\
Max gradient norm & 1.0 \\
Precision & BF16 \\
\bottomrule
\end{tabular}
\caption{Fine-tuning hyperparameters.}
\label{tab:hyperparams}
\end{table}

\paragraph{Evaluation.} During evaluation on Terminal-Bench 2.0, the fine-tuned models are served via vLLM \cite{kwon2023efficient} in non-thinking mode with temperature 0.7, top-p 0.8, top-k 20, and presence penalty 1.5. The agent interacts with the environment through a single headless terminal providing only JSON-formatted tool calls without additional harness or auxiliary tools. The context window token budget is measured using the Qwen3 tokenizer, with a maximum of 37,888 input tokens and 3,072 output tokens per turn.

\section{Diversity Control Examples}
\label{app:categories}

We present representative examples verbatim from our diversity control dimensions to illustrate the design philosophy behind each one.

\subsection{Category Dimension}

Table~\ref{tab:category_list} lists all 39 predefined categories. Below we show one representative category specification in full as used in our framework.
\begin{table}[h]
\centering
\scriptsize
\setlength{\tabcolsep}{8pt}
\begin{tabular}{@{}rl|rl@{}}
\toprule
\textbf{\#} & \textbf{Category} & \textbf{\#} & \textbf{Category} \\
\midrule
1 & Algorithm Design & 21 & Git Operations \\
2 & API Design & 22 & Incident Response \\
3 & Bioinformatics & 23 & Message Queues \\
4 & Code Audit & 24 & ML Training \\
5 & Code Golf & 25 & Monitoring \\
6 & Code Migration & 26 & Networking \\
7 & Compilation \& Build & 27 & Optimization \\
8 & Cryptanalysis & 28 & Physics \& Simulation \\
9 & Database Systems & 29 & Puzzle \& Investigation \\
10 & Data Processing & 30 & Scientific Computing \\
11 & Debugging & 31 & Security \& CTF \\
12 & DevOps \& CI/CD & 32 & Shell Scripting \\
13 & Distributed ML & 33 & Software Engineering \\
14 & Distributed Systems & 34 & System Administration \\
15 & Env. \& Dependencies & 35 & System Orchestration \\
16 & File Operations & 36 & Testing \& QA \\
17 & Financial Engineering & 37 & Text Processing \\
18 & FPGA \& Hardware & 38 & Video \& Multimedia \\
19 & Functional Languages & 39 & Web Development \\
20 & Games \& Simulations & & \\
\bottomrule
\end{tabular}
\caption{Complete list of 39 category dimensions.}
\label{tab:category_list}
\end{table}

\begin{tcolorbox}[colback=gray!5, colframe=black!70, colbacktitle=gray!28, title={\textcolor{orange!70!black}{\textbf{Category: Database and SQL}}}, fonttitle=\small\bfseries, left=4pt, right=4pt, top=4pt, bottom=4pt, boxrule=0.6pt, breakable]
\scriptsize

Tasks involving database operations, SQL queries, schema design, query optimization, and database administration.

\textbf{The examples below are for inspiration only. Be creative and design your own unique task!}

\vspace{3pt}
\textbf{Task Pattern Ideas:}
\begin{itemize}[nosep, leftmargin=*, itemsep=1pt]
\item Write and optimize complex SQL queries
\item Design database schemas for specific requirements
\item Implement database migrations and schema evolution
\item Implement stored procedures or triggers
\item Cross-database data migration (MySQL to PostgreSQL, etc.)
\item Implement full-text search functionality
\item Time-series data management and queries
\item Implement database connection pooling
\item Data integrity validation and cleanup
\item Implement caching strategies with database backends
\end{itemize}

\vspace{3pt}
\textbf{Skills \& Tools} (tag suggestions): Core: \texttt{database}, \texttt{sql}, \texttt{data-management}, \texttt{query-optimization}; Databases: \texttt{sqlite}, \texttt{postgresql}, \texttt{mysql}, \texttt{mongodb}, \texttt{redis}; Languages: \texttt{sql}; Tools: \texttt{sqlite3}, \texttt{psql}, \texttt{mysql-client}, \texttt{pgcli}

\vspace{3pt}
\textbf{Possible Input Data} (be creative with environment files!): SQLite database files (\texttt{.sqlite}, \texttt{.db}); SQL dump files (\texttt{.sql}) with schema and data; Slow query logs for optimization tasks; Schema definition files (DDL scripts); CSV/JSON data for import tasks; Database configuration files; Index statistics and query plans

\vspace{3pt}
\textbf{Verification Approaches:} Verify query results match expected output; Check query performance (execution time, explain plan); Validate schema correctness (constraints, relationships); Test data integrity after migrations; Verify index usage in query plans; Check transaction handling (ACID compliance)

\vspace{3pt}
\textbf{Remember}: These are just starting points. Create database challenges that test SQL proficiency, query optimization skills, and understanding of database internals!
\end{tcolorbox}

\subsection{Scenario Dimension}

Table~\ref{tab:scenario_list} lists all 10 scenario styles. Below we show one representative scenario specification verbatim.

\begin{table}[h]
\centering
\scriptsize
\setlength{\tabcolsep}{8pt}
\begin{tabular}{@{}lp{5.2cm}@{}}
\toprule
\textbf{Scenario} & \textbf{Description} \\
\midrule
Minimal & Ultra-brief 1--3 sentences \\
Structured & Well-organized request with goal and context \\
Narrative & Workplace story with characters \\
Emergency & Urgent time-pressure style \\
Creative & Unusual or playful framing \\
Casual & Informal conversational tone \\
Follow-up & Continuation with new requirements \\
Multi-request & Multiple sub-tasks bundled \\
Specification & Formal specification style \\
Terminal Dump & Starts with pasted terminal output \\
\bottomrule
\end{tabular}
\caption{Complete list of 10 scenario styles.}
\label{tab:scenario_list}
\end{table}

\begin{tcolorbox}[colback=gray!5, colframe=black!70, colbacktitle=gray!28, title={\textcolor{green!50!black}{\textbf{Scenario: Narrative}}}, fonttitle=\small\bfseries, left=4pt, right=4pt, top=4pt, bottom=4pt, boxrule=0.6pt, breakable]
\scriptsize

For the \texttt{instruction.md} you generate, use a \textbf{scenario-based narrative} style:

\vspace{3pt}
\textbf{Key principles:}
\begin{itemize}[nosep, leftmargin=*, itemsep=1pt]
\item \textbf{Tell a story} -- Set up a realistic workplace situation
\item \textbf{Use characters} -- ``Your colleague Marcus'', ``The intern'', ``Sarah from the ops team''
\item \textbf{Explain the situation} -- What's happening and what needs to be done
\item \textbf{Natural flow} -- Don't use formal sections, write as connected paragraphs
\item \textbf{Embedded requirements} -- Weave technical details into the narrative
\end{itemize}

\vspace{3pt}
\textbf{Good example:}

\texttt{The data science team has been collecting sensor readings from 200 IoT devices across three warehouses. The raw data lives in /app/data/sensors/ -- one Parquet file per day, going back 90 days. Lisa from the analytics team needs a consolidated dashboard-ready dataset by end of week. The problem is that the devices use three different firmware versions, each with slightly different data schemas (some use Celsius, others Fahrenheit; timestamp formats vary). Your job: build a pipeline that normalizes all the data into a consistent schema, identifies anomalous readings, and produces a clean daily-aggregated dataset at /app/output/dashboard.parquet.}

\vspace{3pt}
\textbf{Remember}: Make it feel like a real workplace situation -- not a formal specification, but a story with a clear deliverable.
\end{tcolorbox}

\subsection{Difficulty Dimension}

We define four difficulty levels as shown in Table~\ref{tab:difficulty_summary}. In practice, we primarily use the \textit{hard} and \textit{extreme} levels for task generation. Below we show the \textit{hard} level specification verbatim.

\begin{table}[h]
\centering
\scriptsize
\begin{tabular}{@{}lp{2.8cm}p{3.2cm}@{}}
\toprule
\textbf{Level} & \textbf{Skill Scope} & \textbf{Solution Characteristics} \\
\midrule
Easy & Single core tool/technique & Clear input$\rightarrow$process$\rightarrow$output workflow \\
Medium & 2--3 tools combined with judgment calls & Multi-step; agent must choose approach based on discovery \\
Hard & Specialized domain expertise beyond basic programming & Non-obvious path; difficulty from real materials, not artificial constraints \\
Extreme & Deep knowledge requiring active research of provided materials & Discovery-driven; 3+ chained non-obvious insights \\
\bottomrule
\end{tabular}
\caption{Summary of difficulty level criteria.}
\label{tab:difficulty_summary}
\end{table}

\begin{tcolorbox}[colback=gray!5, colframe=black!70, colbacktitle=gray!28, title={\textcolor{blue!60!black}{\textbf{Difficulty: Hard}}}, fonttitle=\small\bfseries, left=4pt, right=4pt, top=4pt, bottom=4pt, boxrule=0.6pt, breakable]
\scriptsize

Your task MUST meet ALL of the following criteria to qualify as HARD difficulty:

\vspace{3pt}
\textbf{1. Specialized Knowledge Required:} The task must require domain expertise that goes beyond basic programming (e.g., cryptographic algorithms, compiler internals, signal processing, bioinformatics). A general-purpose developer should NOT be able to solve it without learning new concepts.

\vspace{2pt}
\textbf{2. Non-Obvious Solution Path:} The solution must NOT be a straightforward ``read input $\rightarrow$ process $\rightarrow$ write output'' pipeline. Require creative problem-solving, algorithm design, or reverse engineering. Include at least one step where the solver must discover or infer something not explicitly stated.

\vspace{2pt}
\textbf{3. Difficulty from the Materials, Not Artificial Constraints:} The challenge should come from the inherent complexity of the real materials. Do NOT make tasks hard by just adding arbitrary time limits or format restrictions on top of a simple problem.

\vspace{2pt}
\textbf{4. Multi-Stage Verification:} Tests must verify multiple intermediate or final artifacts. Include at least 3 distinct verification checks. Verify both correctness AND constraints (performance, size, format).

\vspace{2pt}
\textbf{5. Real-World Complexity:} Model realistic scenarios with edge cases and error conditions. Include at least one ``trap'' that a naive implementation would fail on. The task should take an expert 30+ minutes to solve.

\vspace{3pt}
\textbf{Anti-patterns to AVOID:} Simple CRUD operations; Tasks solvable with a single library call; Problems with obvious, linear solutions; Basic ML training (MNIST-level tasks); Making a simple task ``hard'' by adding artificial constraints.
\end{tcolorbox}

\section{Detailed Error Analysis}
\label{app:error}

To understand the remaining failure modes of Meta-Task-32B, we manually inspect failed trajectories from our evaluation runs on Terminal-Bench 2.0. We exclude infrastructure errors (RuntimeError) and wall-clock timeouts (which reflect API inference latency rather than model decisions), focusing on failures attributable to model behavior. We identify four distinct error categories as follows.

\paragraph{Context Compaction Loss (47.3\%).} When the trajectory exceeds the 37,888-token context budget, the scaffold compacts earlier observations into summaries, causing the agent to lose critical task constraints or error history. The agent then submits solutions that violate forgotten specifications with high confidence. This is strongly correlated with compaction frequency: success rate drops from 54\% (no compaction) to 31\% (1--2 compactions) to 14\% (6+ compactions).

\paragraph{Flawed Self-Verification (32.9\%).} The agent runs ad-hoc inline checks (e.g., \texttt{python3 -c} assertions, manual output inspection) that pass on simple cases, but misses edge cases caught by the actual test suite. The agent then marks the task as complete with false confidence.

\paragraph{Step Budget Exhaustion (15.1\%).} The agent explores multiple strategies without converging on a correct solution before reaching the 1,000-step turn limit. This occurs primarily in tasks requiring specialized domain knowledge where initial approaches are fundamentally misguided.

\paragraph{Repetitive Strategy Loop (4.8\%).} The agent repeats the same failing approach with only minor variations, unable to recognize the need for a fundamentally different strategy. Unlike budget exhaustion, these agents do not explore alternatives at all.

\vspace{4pt}
Table~\ref{tab:error_cases} presents one representative failure for each category, extracted from our actual evaluation trajectories with exact step numbers and verifier outputs.

\begin{table*}[t]
\centering
\small
\renewcommand{\arraystretch}{1.25}
\begin{tabular}{@{}p{3.5cm}p{6.2cm}p{5.5cm}@{}}
\toprule
\textbf{Error Category / Task} & \textbf{Trajectory Summary} & \textbf{Verifier Output \& Root Cause} \\
\midrule
\textbf{Context Compaction Loss} \newline \texttt{dna-assembly} \newline {\scriptsize 221 steps, 11 compactions}
& Agent reads primer design task (Step 2), iteratively designs primers and computes melting temperatures with \texttt{oligotm} (Steps 3--198). After 11 context compactions, marks complete (Step 221) and submits primers containing \texttt{GTTTAAAC\textbf{NNN}GGTCTCGTTAT...}
& \textbf{Verifier:} \texttt{Primer must contain only A, T, C, G.} \newline \textbf{Cause:} Agent retains BsaI site logic (using \texttt{NNN} spacer) but loses the ATCG-only constraint after repeated compaction. \\
\midrule
\textbf{Flawed Self-Verification} \newline \texttt{filter-js-from-html} \newline {\scriptsize 23 steps, 0 compactions}
& Agent builds an HTML XSS filter (Steps 8--18) and tests inline with 5 self-chosen vectors (\texttt{<script>}, \texttt{onclick}, \texttt{javascript:} href). All pass. Marks complete at Step 23: ``Filter thoroughly tested and working.''
& \textbf{Verifier:} \texttt{assert 4 == 0} (4 vector groups failed). \newline \textbf{Cause:} Actual suite applies 400+ vectors with encoded JS, uncommon handlers (\texttt{onscroll}, \texttt{ondataavailable}), and nested obfuscation. \\
\midrule
\textbf{Step Budget Exhaustion} \newline \texttt{polyglot-rust-c} \newline {\scriptsize 1000 steps, 32 compactions}
& Agent attempts a file valid in both Rust and C++ (Steps 1--1000). Oscillates: ``Rust succeeded'' (55$\times$) vs. ``C++ sees invalid code'' (277$\times$). Never marks complete.
& \textbf{Verifier:} Never reached (step limit). \newline \textbf{Cause:} Two languages have conflicting syntax. Agent tries 15+ structural approaches but cannot satisfy both compilers simultaneously. \\
\midrule
\textbf{Repetitive Strategy Loop} \newline \texttt{write-compressor} \newline {\scriptsize 1000 steps, 7 compactions}
& Agent works on compression task (Steps 1--194), then gets trapped in a terminal pager. Sends identical \texttt{C-cq} escape sequence for 699 consecutive steps (Steps 199--1000).
& \textbf{Verifier:} Never reached (step limit). \newline \textbf{Cause:} Correctly identifies stuck state but cycles the same keystrokes indefinitely, never trying \texttt{:q!} or spawning a new shell. \\
\bottomrule
\end{tabular}
\caption{Representative failure cases of Meta-Task-32B on Terminal-Bench 2.0. All step numbers and verifier messages are from actual evaluation runs.}
\label{tab:error_cases}
\end{table*}

\input{sections/appendix_task_review}

\section{Synthesized Task Example}
\label{app:task_examples}

\begin{figure*}[h]
  \centering
  \includegraphics[width=\textwidth]{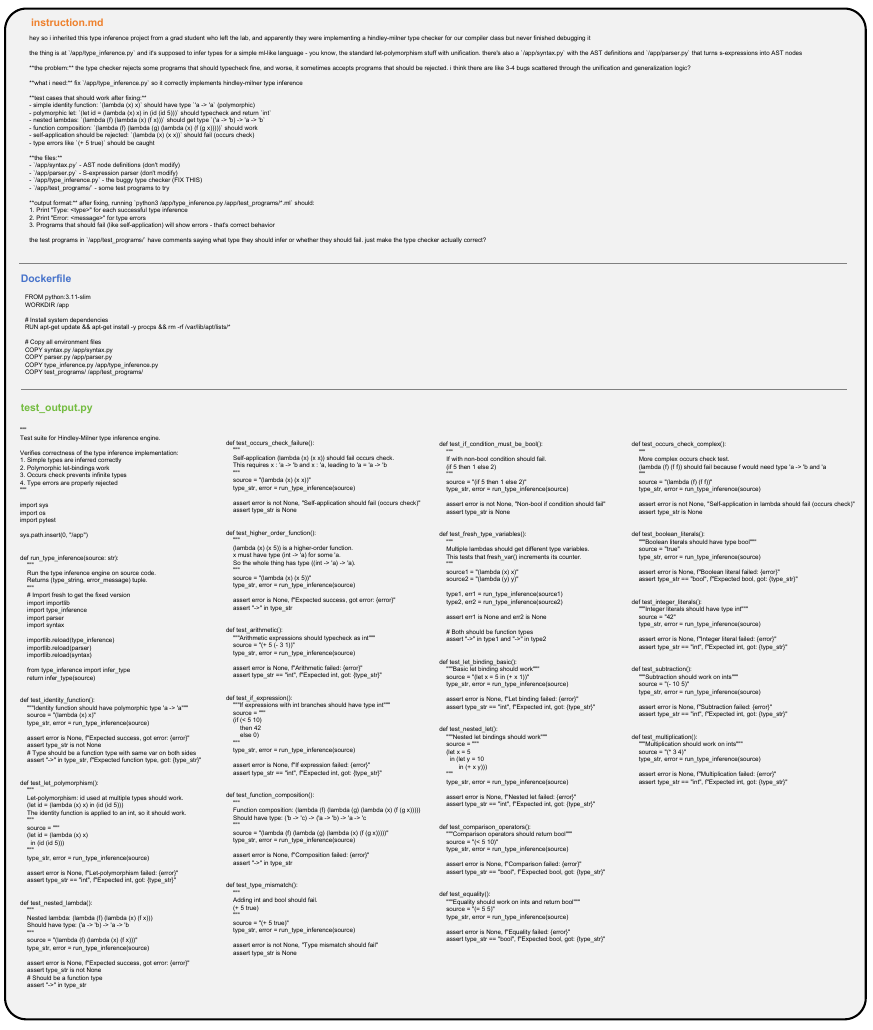}
  \caption{A synthesized task example from Meta-Task. Top: the task instruction describing a buggy type checker to fix. Middle: the Dockerfile setting up the environment. Bottom: the pytest verification tests.}
  \label{fig:task_case}
\end{figure*}

Figure~\ref{fig:task_case} presents a representative synthesized task (Hindley-Milner type inference debugging, hard difficulty), showing the three core components of a task package from top to bottom: the task instruction, the environment Dockerfile, and the verification tests.

\input{sections/appendix_prompts}

%% file: sections/appendix_task_review.tex
\section{Task Quality Evaluation Details}
\label{app:task_quality}

We employ Claude Opus 4.6 as an automated reviewer following the official Terminal-Bench review pipeline. For each synthesis method, we randomly sample 50 tasks and conduct two types of review:

\textbf{Proposal Review.} The reviewer reads the task description and evaluates whether the task idea is likely to be accepted into Terminal-Bench based on the official Task Proposal Rubric, which requires tasks to be verifiable, well-specified, solvable, appropriately challenging, realistic, and outcome-verified. The reviewer outputs Accept, Uncertain, or Reject.

\textbf{Implementation Review.} The reviewer reads the full task package (instruction, Dockerfile, solution, tests, task.toml) and evaluates it against the 19-criterion Task Implementation Rubric. The 19 criteria cover: verifiable, well\_specified, solvable, difficult, novel, anti\_cheat\_robustness, interesting, agentic, functional\_verification, outcome\_verified, deterministic\_reproducible, essential\_difficulty, instruction\_clarity, solution\_quality, environment\_hygiene, test\_instruction\_alignment, reviewable, structured\_data\_schema, and typos. Each criterion is scored as PASS, FAIL, or NA.

The implementation compliance rate reported in Figure~\ref{fig:task_analysis}(a) is the average PASS rate across all 19 criteria, reflecting the overall completeness and reliability of the synthesized task package.

%% file: sections/appendix_prompts.tex
\section{Prompt Templates}
\label{app:prompts}

We present the key prompts used in the Meta-Task pipeline. The variable parts (category, scenario, difficulty) are illustrated in Appendix~\ref{app:categories}.

\subsection{Meta-Task Instruction Template}

The meta-task instruction consists of fixed parts (defining the workflow, output structure, and validation rules) and variable parts (category, scenario, difficulty). Figure~\ref{fig:prompt_metatask} shows the fixed parts with variable placeholders marked as \texttt{[...]}.

\subsection{Multi-Phase Synthesis Prompt}

In multi-phase mode (\S\ref{sec:multiphase}), a lightweight Phase 1 meta-task instructs an agent to dynamically design a new (category, scenario) specification pair (Figure~\ref{fig:prompt_phase1}). The generated specifications are then injected into the standard meta-task instruction for Phase 2 synthesis.

\subsection{Trajectory-Level LLM-as-Judge Prompt}

The LLM-as-Judge reviews successful trajectories (Figure~\ref{fig:prompt_judge}) to determine whether they demonstrate genuine problem-solving behavior suitable for SFT training.

\onecolumn  

\begin{tcolorbox}[colback=gray!5, colframe=black!70, colbacktitle=gray!28, title={\textcolor{blue!60!black}{\textbf{Meta-Task Instruction (Fixed Parts)}}}, fonttitle=\small\bfseries, left=6pt, right=6pt, top=6pt, bottom=6pt, boxrule=0.6pt, breakable]
\small

\textbf{\# Task: Generate a Terminal-Based AI Agent Task}

You are an expert in \texttt{[CATEGORY]} with deep knowledge of best practices, common patterns, and real-world challenges in this domain. Your task is to create a high-quality terminal-based task that tests an AI agent's ability to solve practical problems in a containerized environment.

\vspace{6pt}
\textbf{\#\# 1. Reference}

Study the following examples in \texttt{/app/examples/}: \texttt{[EXAMPLE\_LIST]}. Start by exploring these. The real example shows the expected format, structure, and quality level. \textbf{Do not imitate the example's content, topic, or problem design.} Create something completely original.

\vspace{6pt}
\textbf{\#\# 2. System Architecture}

Your generated task is a self-contained package used in three stages:
\begin{itemize}[nosep, leftmargin=*, itemsep=2pt]
\item \textbf{Synthesis Environment (current):} You create task files in \texttt{/app/output/}. You can run \texttt{solve.sh} and tests to validate your task. You do NOT have Docker permissions here.
\item \textbf{Agent Execution Environment:} Built using \texttt{docker build -f environment/Dockerfile}. The solving agent works here to complete \texttt{instruction.md}. Files from \texttt{environment/} are COPYed into the container. Agent has full terminal access but limited time.
\item \textbf{Test Verification:} \texttt{tests/} directory is mounted at \texttt{/tests/}. \texttt{test.sh} runs \texttt{pytest /tests/test\_outputs.py} and writes 1 (pass) or 0 (fail) to \texttt{/logs/verifier/reward.txt}.
\end{itemize}

\vspace{2pt}
\textbf{The task must be fully self-contained.} A fresh environment is built from your Dockerfile alone. All input data, dependencies, config files, and code referenced in \texttt{instruction.md} must be present inside the built container — either COPYed from \texttt{environment/} or generated during \texttt{docker build}.

\vspace{6pt}
\textbf{\#\# 3. Output Structure} — Generate a \texttt{[DIFFICULTY]} task in \texttt{/app/output/}:
\begin{verbatim}
/app/output/
|-- instruction.md          # Task description (absolute paths)
|-- task.toml               # Metadata (difficulty, category, tags)
|-- environment/
|   |-- Dockerfile          # FROM, WORKDIR, deps, data generation
|   +-- [data/scripts/...]  # (Optional) Input files, configs
|-- solution/
|   +-- solve.sh            # Reference solution (not shown to agent)
+-- tests/
    |-- test.sh             # Runs pytest, writes reward
    +-- test_outputs.py     # Pytest verification tests
\end{verbatim}

\vspace{-2pt}
\textbf{File-by-file guidance:}
\begin{itemize}[nosep, leftmargin=*, itemsep=1pt]
\item \texttt{instruction.md} — Task description; state the goal, not the solution; use absolute paths
\item \texttt{task.toml} — Fill in difficulty, category, and tags
\item \texttt{environment/Dockerfile} — All input data must be present after \texttt{docker build}
\item \texttt{solution/solve.sh} — Reference solution proving solvability (hidden from solving agent)
\item \texttt{tests/test\_outputs.py} — Pytest tests that verify the solution
\end{itemize}

\vspace{6pt}
\textbf{\#\# 4. Task Category:} \texttt{[CATEGORY\_CONTENT]}

\vspace{6pt}
\textbf{\#\# 5. Writing Style:} \texttt{[SCENARIO\_CONTENT]}

Critical rules for \texttt{instruction.md}:
\begin{itemize}[nosep, leftmargin=*, itemsep=1pt]
\item \textbf{Describe the goal, not the solution.} State what the agent should produce or achieve. Don't explain the approach or give step-by-step guidance.
\item \textbf{Keep it concise.} Match the scenario's detail level. Look at the real examples for calibration.
\item \textbf{Use light formatting.} Plain text and simple bullet points. The instruction should read naturally.
\item \textbf{Let the environment carry detail.} Put complexity in environment files, not in the instruction text. The agent should explore to understand the full picture.
\end{itemize}

\vspace{6pt}
\textbf{\#\# 6. Difficulty Requirements:} \texttt{[DIFFICULTY\_SECTION]}

\vspace{6pt}
\textbf{\#\# 7. Dockerfile Best Practices}
\begin{itemize}[nosep, leftmargin=*, itemsep=1pt]
\item \textbf{NEVER use heredoc syntax} (\texttt{<< EOF}) — causes Docker build failure. Use \texttt{printf}, \texttt{echo}, \texttt{python3 -c}, or COPY a file instead.
\item \textbf{NEVER COPY solution/ or tests/ into the image} — they are mounted at runtime.
\item \textbf{Verify every COPY source} exists under \texttt{environment/} before finalizing.
\item \textbf{Pin dependency versions} when stability matters: \texttt{pip install numpy==1.24.0}
\item \textbf{Delete setup scripts after execution}: \texttt{RUN bash setup.sh \&\& rm setup.sh}
\item \textbf{Required base}: \texttt{FROM python:3.11-slim}, \texttt{WORKDIR /app}, install \texttt{procps}
\end{itemize}

\vspace{6pt}
\textbf{\#\# 8. Task Construction Approaches}
\begin{itemize}[nosep, leftmargin=*, itemsep=1pt]
\item \textbf{Implement from scratch} — Agent builds something new given a goal
\item \textbf{Reverse engineer} — Agent figures out what a binary, data file, or system does
\item \textbf{Build / compile} — Agent compiles a real project from source, resolving dependencies
\item \textbf{Configure / deploy} — Agent sets up a system or service
\item \textbf{Analyze / extract} — Agent analyzes data or files to produce derived output
\item \textbf{Optimize} — Working code provided, agent must make it faster/smaller
\item \textbf{Migrate / convert} — Agent converts code or data between formats/languages/versions
\item \textbf{Debug / fix} — Broken code provided, agent must find and fix the issue
\end{itemize}

\vspace{6pt}
\textbf{\#\# 9. Self-Validation (REQUIRED)}

Before finishing, you MUST validate your task is solvable and consistent:
\begin{enumerate}[nosep, leftmargin=*, itemsep=2pt]
\item \textbf{Execute your solution:} \texttt{bash /app/output/solution/solve.sh}
\item \textbf{Run tests to verify:} \texttt{pytest /app/output/tests/test\_outputs.py -v}
\item \textbf{Verify consistency across all components:}
\begin{itemize}[nosep, leftmargin=*, itemsep=1pt]
\item \textbf{instruction $\leftrightarrow$ solution:} Does \texttt{solve.sh} actually accomplish what \texttt{instruction.md} asks?
\item \textbf{instruction $\leftrightarrow$ environment:} Does \texttt{Dockerfile} provide all files/tools mentioned in instruction?
\item \textbf{solution $\leftrightarrow$ tests:} Does \texttt{solve.sh} output match exactly what tests expect (paths, formats, values)?
\item \textbf{tests $\leftrightarrow$ instruction:} Every test traces back to a requirement in \texttt{instruction.md}, and every requirement has a corresponding test.
\item \textbf{No answer leakage:} Re-read \texttt{instruction.md} as if you're the solving agent — does it reveal the solution approach?
\item \textbf{Tests are robust:} Tests verify behavior through execution (run code, check results), not just string matching or grep.
\end{itemize}
\item \textbf{Verify difficulty requirements} — Re-read the Difficulty section and verify your task meets each criterion.
\item \textbf{If any check fails, fix and repeat from Step 1.}
\item \textbf{Clean up} — Remove any output files created during validation.
\end{enumerate}

\vspace{6pt}
\textbf{\#\# 10. Final Checklist}
\begin{itemize}[nosep, leftmargin=*, itemsep=1pt]
\item All 6 required files present in \texttt{/app/output/}
\item Solvability: \texttt{solve.sh} runs successfully and produces correct output
\item Tests pass: \texttt{pytest test\_outputs.py -v} shows all tests green
\item Tests are general: any correct solution should pass, not just your specific \texttt{solve.sh}
\item Output discoverable: the solving agent can figure out what to produce from \texttt{instruction.md}
\item Self-contained: all paths, data, and dependencies exist inside the built container
\item Consistency: instruction $\leftrightarrow$ solution $\leftrightarrow$ tests $\leftrightarrow$ environment all correctly aligned
\item Originality: task is genuinely different from examples
\item Dockerfile: no heredoc syntax, valid packages, all COPY sources exist
\item No answer leakage: \texttt{instruction.md} states the goal without revealing the solution approach
\item Instruction style: follows the Writing Style rules — concise, natural, scenario-appropriate
\item Difficulty: task genuinely meets \texttt{[DIFFICULTY]} criteria
\end{itemize}

\end{tcolorbox}
\captionof{figure}{Meta-Task synthesis instruction template (fixed parts). Variable placeholders \texttt{[...]} are filled with category, scenario, and difficulty specifications at each generation.}
\label{fig:prompt_metatask}

\vspace{12pt}

\begin{tcolorbox}[colback=gray!5, colframe=black!70, colbacktitle=gray!28, title={\textcolor{purple!60!black}{\textbf{Phase 1: Generate a New Category and Scenario}}}, fonttitle=\small\bfseries, left=6pt, right=6pt, top=6pt, bottom=6pt, boxrule=0.6pt, breakable]
\small

You are a creative task designer. Your job is to generate a \textbf{category template} and a \textbf{scenario template} that will guide the creation of a terminal-based evaluation task where an AI agent works in a Linux Docker container, using the terminal to accomplish goals through code, commands, and tools.

These tasks test an agent's ability to \textbf{plan, explore, and accomplish goals}. Given a goal (sometimes vague), the agent must figure out how to achieve it by exploring the environment and making decisions on its own.

\textbf{Target difficulty: \texttt{\$difficulty}.}

\texttt{\$difficulty\_requirements}

Your task patterns and examples should reflect this difficulty level. Your output is a starting point that another agent will build on, so aim for diversity and creativity.

\vspace{6pt}
\textbf{Your Topic: \texttt{\$topic}}

Generate a category template around this topic. Below is a reference category template (on a different topic) showing the format and level of detail to aim for:

\texttt{\$category\_ref}

\vspace{6pt}
\textbf{Reference Scenario Template}

Below is an example scenario template showing the format. Generate a \textbf{new} scenario template with a different style.

\texttt{\$scenario\_ref}

\vspace{6pt}
\textbf{Your Task}

Write your output to the file \texttt{/app/phase1\_output.md} containing a new category (around the topic \texttt{\$topic}) and a new scenario.

\textbf{For the category:} Build around \texttt{\$topic}. Come up with diverse task patterns, tools, data types, and challenges specific to this topic. Use the reference above only for format guidance.

\textbf{For the scenario:} Your scenario must follow this constraint:

\texttt{\$scenario\_constraint}

Design a scenario template that produces \texttt{instruction.md} files matching the constraint above. The scenario should specify the writing style, detail level, tone, and information strategy.

\vspace{4pt}
\textbf{Output format:}
\begin{verbatim}
=== CATEGORY ===
### Category: [Name]
[Your category content]

=== SCENARIO ===
For the `instruction.md` you generate, use a
**[style name]** style:
[Your scenario content]
\end{verbatim}

\end{tcolorbox}
\captionof{figure}{Multi-phase synthesis prompt (Phase 1). The agent dynamically designs a new category and scenario specification from seed signals.}
\label{fig:prompt_phase1}

\vspace{12pt}

\begin{tcolorbox}[colback=gray!5, colframe=black!70, colbacktitle=gray!28, title={\textcolor{teal!60!black}{\textbf{Trajectory Quality Review Prompt}}}, fonttitle=\small\bfseries, left=6pt, right=6pt, top=6pt, bottom=6pt, boxrule=0.6pt, breakable]
\small

\textbf{System:} You are a training data quality reviewer for AI agent trajectories. We are training an AI agent (via SFT) to solve terminal-based tasks in Docker containers. Each trajectory is a sequence of (observation, action) pairs. We want to keep only trajectories that teach GOOD problem-solving behavior. The core question: \textbf{Does this trajectory demonstrate the kind of behavior we want the trained model to replicate?}

\vspace{6pt}
\textbf{What Makes a GOOD Trajectory (keep):}
\begin{itemize}[nosep, leftmargin=*, itemsep=2pt]
\item Understanding a problem through exploration and reasoning
\item Implementing solutions by writing its own code
\item Testing, diagnosing failures, and iterating with meaningful changes
\item Making progress through its own capability, not through information shortcuts
\end{itemize}

\vspace{6pt}
\textbf{What Makes a BAD Trajectory (discard):}

\vspace{2pt}
\textit{Shortcutting (agent avoids doing the actual work):}
\begin{itemize}[nosep, leftmargin=*, itemsep=2pt]
\item Agent obtains answers from the environment rather than computing them — reading files that contain pre-computed results, answers, hints, or solution algorithms
\item Agent reads files with red-flag names (``solution'', ``fixed'', ``answer'', ``expected\_output'', ``correct'', ``bugfix'', ``hint'', ``cheat'')
\item Agent runs pre-existing scripts that produce the required output rather than writing its own implementation
\item Agent sees computed values in data files and directly uses them as output without independently deriving them
\item Agent reads source code comments that explicitly state the bug and correct fix (e.g., ``BUG: should be X instead of Y'')
\end{itemize}

\vspace{2pt}
\textit{Fabrication (agent produces fake results):}
\begin{itemize}[nosep, leftmargin=*, itemsep=2pt]
\item Agent claims to have verified or computed something but terminal shows it never ran
\item Agent produces specific numerical values with no derivation trace anywhere in the trajectory
\item Agent marks task complete despite terminal being stuck or non-functional
\item Agent writes ``verification passed'' when the actual test command failed or was never executed
\end{itemize}

\vspace{2pt}
\textit{Unproductive behavior (low learning signal):}
\begin{itemize}[nosep, leftmargin=*, itemsep=2pt]
\item Agent repeats the same approach many times without meaningful adaptation
\item Agent sends $>$10 consecutive turns of empty commands, Ctrl+C/Ctrl+D without recovery
\item Agent tries the same general approach $>$15 times without strategy changes
\item Trajectory is too short to contain substantive work
\end{itemize}

\vspace{6pt}
\textbf{Key Distinction:} Reading input data that the agent needs to PROCESS (log files, databases, CSV data) is NOT contamination. Reading reference documentation (API docs, library source code) is NOT contamination. The key test: does the file contain the ANSWER to what the agent is supposed to produce, or does it contain RAW DATA that the agent must still reason about?

\vspace{6pt}
\textbf{Output:} KEEP or DISCARD with brief justification.
\end{tcolorbox}
\captionof{figure}{Trajectory-level LLM-as-Judge quality review prompt for filtering SFT training data.}
\label{fig:prompt_judge}

%% file: acl_latex.bbl
\begin{thebibliography}{35}
\providecommand{\natexlab}[1]{#1}

\bibitem[{Anthropic(2025)}]{anthropic2025claudecode}
Anthropic. 2025.
\newblock Claude code: Best practices for agentic coding.
\newblock \url{https://www.anthropic.com/engineering/claude-code-best-practices}.

\bibitem[{Bui(2026)}]{bui2026building}
Nghi~DQ Bui. 2026.
\newblock Building effective ai coding agents for the terminal: Scaffolding, harness, context engineering, and lessons learned.
\newblock \emph{arXiv preprint arXiv:2603.05344}.

\bibitem[{Chen et~al.(2026)Chen, Zhang, Feng, Wang, Zhao, Cao, Yang, Chen, Li, Ma, Ge, Zhang, Cui, Liu, Zhou, Sun, Lin, and Hui}]{chen2026sweuniv}
Mouxiang Chen, Lei Zhang, Yunlong Feng, Xuwu Wang, Wenting Zhao, Ruisheng Cao, Jiaxi Yang, Jiawei Chen, Mingze Li, Zeyao Ma, Hao Ge, Zongmeng Zhang, Zeyu Cui, Dayiheng Liu, Jingren Zhou, Jianling Sun, Junyang Lin, and Binyuan Hui. 2026.
\newblock \href {https://arxiv.org/abs/2602.02361} {Swe-universe: Scale real-world verifiable environments to millions}.
\newblock \emph{Preprint}, arXiv:2602.02361.

\bibitem[{Cheng et~al.(2026)Cheng, Huang, Gu, Song, Chen, Dong, Zhao, Wen, and Wei}]{cheng2026llm}
Daixuan Cheng, Shaohan Huang, Yuxian Gu, Huatong Song, Guoxin Chen, Li~Dong, Wayne~Xin Zhao, Ji-Rong Wen, and Furu Wei. 2026.
\newblock Llm-in-sandbox elicits general agentic intelligence.
\newblock \emph{arXiv preprint arXiv:2601.16206}.

\bibitem[{Dong et~al.(2026)Dong, Lu, Huang, Zhong, Liu, Huang, Li, Zhao, Song, Li et~al.}]{dong2026agent}
Guanting Dong, Junting Lu, Junjie Huang, Wanjun Zhong, Longxiang Liu, Shijue Huang, Zhenyu Li, Yang Zhao, Xiaoshuai Song, Xiaoxi Li, and 1 others. 2026.
\newblock Agent-world: Scaling real-world environment synthesis for evolving general agent intelligence.
\newblock \emph{arXiv preprint arXiv:2604.18292}.

\bibitem[{Fan et~al.(2026)Fan, Yu, Cai, Guan, Yang, Hu, Zhou, Wu, Han, Zhang et~al.}]{fan2026toward}
Zhiyuan Fan, Tinghao Yu, Yuanjun Cai, Jiangtao Guan, Yun Yang, Dingxin Hu, Jiang Zhou, Xing Wu, Zhuo Han, Feng Zhang, and 1 others. 2026.
\newblock Toward scalable terminal task synthesis via skill graphs.
\newblock \emph{arXiv preprint arXiv:2604.25727}.

\bibitem[{Gandhi et~al.(2026)Gandhi, Garg, Goodman, and Papailiopoulos}]{gandhi2026endless}
Kanishk Gandhi, Shivam Garg, Noah~D Goodman, and Dimitris Papailiopoulos. 2026.
\newblock Endless terminals: Scaling rl environments for terminal agents.
\newblock \emph{arXiv preprint arXiv:2601.16443}.

\bibitem[{{Harbor Framework Team}(2026)}]{Harbor_Framework}
{Harbor Framework Team}. 2026.
\newblock \href {https://github.com/harbor-framework/harbor} {{Harbor: A framework for evaluating and optimizing agents and models in container environments}}.

\bibitem[{Hu et~al.(2022)Hu, Shen, Wallis, Allen-Zhu, Li, Wang, Wang, Chen et~al.}]{hu2022lora}
Edward~J Hu, Yelong Shen, Phillip Wallis, Zeyuan Allen-Zhu, Yuanzhi Li, Shean Wang, Liang Wang, Weizhu Chen, and 1 others. 2022.
\newblock {LoRA}: Low-rank adaptation of large language models.
\newblock In \emph{ICLR}, volume~1, page~3.

\bibitem[{Jimenez et~al.(2024)Jimenez, Yang, Wettig, Yao, Pei, Press, and Narasimhan}]{jimenez2024swe}
Carlos~E Jimenez, John Yang, Alexander Wettig, Shunyu Yao, Kexin Pei, Ofir Press, and Karthik Narasimhan. 2024.
\newblock Swe-bench: Can language models resolve real-world github issues?
\newblock In \emph{International Conference on Learning Representations}, volume 2024, pages 54107--54157.

\bibitem[{Kwon et~al.(2023)Kwon, Li, Zhuang, Sheng, Zheng, Yu, Gonzalez, Zhang, and Stoica}]{kwon2023efficient}
Woosuk Kwon, Zhuohan Li, Siyuan Zhuang, Ying Sheng, Lianmin Zheng, Cody~Hao Yu, Joseph~E. Gonzalez, Hao Zhang, and Ion Stoica. 2023.
\newblock Efficient memory management for large language model serving with pagedattention.
\newblock In \emph{Proceedings of the ACM SIGOPS 29th Symposium on Operating Systems Principles}.

\bibitem[{Lee et~al.(2026)Lee, Nair, Zhang, Lee, Khattab, and Finn}]{lee2026meta}
Yoonho Lee, Roshen Nair, Qizheng Zhang, Kangwook Lee, Omar Khattab, and Chelsea Finn. 2026.
\newblock Meta-harness: End-to-end optimization of model harnesses.
\newblock \emph{arXiv preprint arXiv:2603.28052}.

\bibitem[{Lin et~al.(2018)Lin, Wang, Zettlemoyer, and Ernst}]{lin2018nl2bash}
Xi~Victoria Lin, Chenglong Wang, Luke Zettlemoyer, and Michael~D. Ernst. 2018.
\newblock \href {https://arxiv.org/abs/1802.08979} {Nl2bash: A corpus and semantic parser for natural language interface to the linux operating system}.
\newblock \emph{Preprint}, arXiv:1802.08979.

\bibitem[{Lin et~al.(2026)Lin, Wang, Wu, Fan, Pan, Zhao, and Tu}]{lin2026cli}
Yusong Lin, Haiyang Wang, Shuzhe Wu, Lue Fan, Feiyang Pan, Sanyuan Zhao, and Dandan Tu. 2026.
\newblock Cli-gym: Scalable cli task generation via agentic environment inversion.
\newblock \emph{arXiv preprint arXiv:2602.10999}.

\bibitem[{Meng et~al.(2026)Meng, Wang, Chen, Wang, Lu, Wu, Gao, Wu, and Hu}]{harness_survey}
Qianyu Meng, Yanan Wang, Liyi Chen, Qimeng Wang, Chengqiang Lu, Wei Wu, Yan Gao, Yi~Wu, and Yao Hu. 2026.
\newblock \href {https://doi.org/10.20944/preprints202604.0428.v2} {Agent harness for large language model agents: A survey}.
\newblock \emph{Preprints}.

\bibitem[{Merrill et~al.(2026)Merrill, Shaw, Carlini, Li, Raj, Bercovich, Shi, Shin, Walshe, Buchanan et~al.}]{merrill2026terminal}
Mike~A Merrill, Alexander~G Shaw, Nicholas Carlini, Boxuan Li, Harsh Raj, Ivan Bercovich, Lin Shi, Jeong~Yeon Shin, Thomas Walshe, E~Kelly Buchanan, and 1 others. 2026.
\newblock Terminal-bench: Benchmarking agents on hard, realistic tasks in command line interfaces.
\newblock \emph{arXiv preprint arXiv:2601.11868}.

\bibitem[{OpenAI(2025)}]{openai2025codex}
OpenAI. 2025.
\newblock Introducing codex.
\newblock \url{https://openai.com/index/introducing-codex/}.

\bibitem[{{OpenClaw}(2026)}]{openclaw2026}
{OpenClaw}. 2026.
\newblock Openclaw.
\newblock \url{https://github.com/openclaw/openclaw}.
\newblock Open-source personal AI assistant, version 2026.3.8, accessed 2026-03-09.

\bibitem[{Pi et~al.(2026)Pi, Lam, Shoeybi, Jannaty, Catanzaro, and Ping}]{pi2026data}
Renjie Pi, Grace Lam, Mohammad Shoeybi, Pooya Jannaty, Bryan Catanzaro, and Wei Ping. 2026.
\newblock On data engineering for scaling llm terminal capabilities.
\newblock \emph{arXiv preprint arXiv:2602.21193}.

\bibitem[{Qin et~al.(2023)Qin, Liang, Ye, Zhu, Yan, Lu, Lin, Cong, Tang, Qian, Zhao, Hong, Tian, Xie, Zhou, Gerstein, Li, Liu, and Sun}]{qin2023toolllm}
Yujia Qin, Shihao Liang, Yining Ye, Kunlun Zhu, Lan Yan, Yaxi Lu, Yankai Lin, Xin Cong, Xiangru Tang, Bill Qian, Sihan Zhao, Lauren Hong, Runchu Tian, Ruobing Xie, Jie Zhou, Mark Gerstein, Dahai Li, Zhiyuan Liu, and Maosong Sun. 2023.
\newblock \href {https://arxiv.org/abs/2307.16789} {Toolllm: Facilitating large language models to master 16000+ real-world apis}.
\newblock \emph{Preprint}, arXiv:2307.16789.

\bibitem[{Ren et~al.(2026)Ren, Wu, Li, Zhu, Xu, Feng, Yuan, Zhang, Batista-Navarro, Yang et~al.}]{ren2026self}
Jincheng Ren, Siwei Wu, Yizhi Li, Kang Zhu, Shu Xu, Boyu Feng, Ruibin Yuan, Wei Zhang, Riza Batista-Navarro, Jian Yang, and 1 others. 2026.
\newblock A self-evolving framework for efficient terminal agents via observational context compression.
\newblock \emph{arXiv preprint arXiv:2604.19572}.

\bibitem[{Shen et~al.(2026)Shen, Rainton, Aliev, Awelkair, Ma, Huang, Mao, Fan, Torr, Ghanem, Hu, Thakker, and Li}]{seta}
Qijia Shen, Jay Rainton, Aznaur Aliev, Ahmed Awelkair, Boyuan Ma, Zhiqi~(Julie) Huang, Yuzhen Mao, Wendong Fan, Philip Torr, Bernard Ghanem, Changran Hu, Urmish Thakker, and Guohao Li. 2026.
\newblock \href {https://github.com/camel-ai/seta} {{SETA: Scaling Environments for Terminal Agents}}.

\bibitem[{Wang et~al.(2025{\natexlab{a}})Wang, Ramalho, Celestino, Pham, Liu, Sinha, Portillo, Osunwa, and Maduekwe}]{wang2025swe}
Lilin Wang, Lucas Ramalho, Alan Celestino, Phuc~Anthony Pham, Yu~Liu, Umang~Kumar Sinha, Andres Portillo, Onassis Osunwa, and Gabriel Maduekwe. 2025{\natexlab{a}}.
\newblock Swe-bench++: A framework for the scalable generation of software engineering benchmarks from open-source repositories.
\newblock \emph{arXiv preprint arXiv:2512.17419}.

\bibitem[{Wang et~al.(2025{\natexlab{b}})Wang, Li, Song, Xu, Tang, Zhuge, Pan, Song, Li, Singh et~al.}]{wang2025openhands}
Xingyao Wang, Boxuan Li, Yufan Song, Frank~F Xu, Xiangru Tang, Mingchen Zhuge, Jiayi Pan, Yueqi Song, Bowen Li, Jaskirat Singh, and 1 others. 2025{\natexlab{b}}.
\newblock Openhands: An open platform for ai software developers as generalist agents.
\newblock In \emph{International Conference on Learning Representations}, volume 2025, pages 65882--65919.

\bibitem[{Wang et~al.(2026)Wang, Chen, Jin, Wang, and Yang}]{wang2026openclaw}
Yinjie Wang, Xuyang Chen, Xiaolong Jin, Mengdi Wang, and Ling Yang. 2026.
\newblock Openclaw-rl: Train any agent simply by talking.
\newblock \emph{arXiv preprint arXiv:2603.10165}.

\bibitem[{Wu et~al.(2026)Wu, Li, Song, Zhang, Wang, Batista-Navarro, Yang, Tang, Dai, Yang et~al.}]{wu2026large}
Siwei Wu, Yizhi Li, Yuyang Song, Wei Zhang, Yang Wang, Riza Batista-Navarro, Xian Yang, Mingjie Tang, Bryan Dai, Jian Yang, and 1 others. 2026.
\newblock Large-scale terminal agentic trajectory generation from dockerized environments.
\newblock \emph{arXiv preprint arXiv:2602.01244}.

\bibitem[{Xia et~al.(2025)Xia, Wang, Yang, Wei, and Zhang}]{xia2025live}
Chunqiu~Steven Xia, Zhe Wang, Yan Yang, Yuxiang Wei, and Lingming Zhang. 2025.
\newblock Live-swe-agent: Can software engineering agents self-evolve on the fly?
\newblock \emph{arXiv preprint arXiv:2511.13646}.

\bibitem[{Yang et~al.(2025)Yang, Li, Yang, Zhang, Hui, Zheng, Yu, Gao, Huang, Lv et~al.}]{yang2025qwen3}
An~Yang, Anfeng Li, Baosong Yang, Beichen Zhang, Binyuan Hui, Bo~Zheng, Bowen Yu, Chang Gao, Chengen Huang, Chenxu Lv, and 1 others. 2025.
\newblock Qwen3 technical report.
\newblock \emph{arXiv preprint arXiv:2505.09388}.

\bibitem[{Yang et~al.(2024)Yang, Jimenez, Wettig, Lieret, Yao, Narasimhan, and Press}]{yang2024sweagent}
John Yang, Carlos~E Jimenez, Alexander Wettig, Kilian Lieret, Shunyu Yao, Karthik Narasimhan, and Ofir Press. 2024.
\newblock {SWE-agent}: Agent-computer interfaces enable automated software engineering.
\newblock \emph{Advances in Neural Information Processing Systems}, 37:50528--50652.

\bibitem[{Yang et~al.(2026)Yang, Lieret, Jimenez, Wettig, Khandpur, Zhang, Hui, Press, Schmidt, and Yang}]{yang2026swe}
John Yang, Kilian Lieret, Carlos Jimenez, Alexander Wettig, Kabir Khandpur, Yanzhe Zhang, Binyuan Hui, Ofir Press, Ludwig Schmidt, and Diyi Yang. 2026.
\newblock Swe-smith: Scaling data for software engineering agents.
\newblock \emph{Advances in Neural Information Processing Systems}, 38.

\bibitem[{Yang et~al.(2023)Yang, Prabhakar, Narasimhan, and Yao}]{yang2023intercode}
John Yang, Akshara Prabhakar, Karthik Narasimhan, and Shunyu Yao. 2023.
\newblock \href {https://arxiv.org/abs/2306.14898} {Intercode: Standardizing and benchmarking interactive coding with execution feedback}.
\newblock \emph{Preprint}, arXiv:2306.14898.

\bibitem[{Yao et~al.(2023)Yao, Zhao, Yu, Du, Shafran, Narasimhan, and Cao}]{yao2023react}
Shunyu Yao, Jeffrey Zhao, Dian Yu, Nan Du, Izhak Shafran, Karthik Narasimhan, and Yuan Cao. 2023.
\newblock \href {https://arxiv.org/abs/2210.03629} {React: Synergizing reasoning and acting in language models}.
\newblock \emph{Preprint}, arXiv:2210.03629.

\bibitem[{Zan et~al.(2025)Zan, Huang, Liu, Chen, Zhang, Xin, Chen, Liu, Zhong, Li et~al.}]{zan2025multi}
Daoguang Zan, Zhirong Huang, Wei Liu, Hanwu Chen, Linhao Zhang, Shulin Xin, Lu~Chen, Qi~Liu, Xiaojian Zhong, Aoyan Li, and 1 others. 2025.
\newblock Multi-swe-bench: A multilingual benchmark for issue resolving.
\newblock \emph{arXiv preprint arXiv:2504.02605}.

\bibitem[{Zhao et~al.(2025)Zhao, Huang, Hu, Wang, Mao, Zhang, Jiang, Wu, Ai, Wang et~al.}]{zhao2025swift}
Yuze Zhao, Jintao Huang, Jinghan Hu, Xingjun Wang, Yunlin Mao, Daoze Zhang, Zeyinzi Jiang, Zhikai Wu, Baole Ai, Ang Wang, and 1 others. 2025.
\newblock Swift: a scalable lightweight infrastructure for fine-tuning.
\newblock In \emph{Proceedings of the AAAI Conference on Artificial Intelligence}, volume~39, pages 29733--29735.

\bibitem[{Zhu et~al.(2026)Zhu, Nie, Li, Huang, Wu, Liu, Sun, Yin, Wang, Liu et~al.}]{zhu2026termigen}
Kaijie Zhu, Yuzhou Nie, Yijiang Li, Yiming Huang, Jialian Wu, Jiang Liu, Ximeng Sun, Zhenfei Yin, Lun Wang, Zicheng Liu, and 1 others. 2026.
\newblock Termigen: High-fidelity environment and robust trajectory synthesis for terminal agents.
\newblock \emph{arXiv preprint arXiv:2602.07274}.

\end{thebibliography}
